\documentclass{article}
\usepackage{spconf,amsmath,graphicx,hyperref}
\usepackage{appendix}
\usepackage{spconf,amsmath,graphicx}
\usepackage{cite}
\usepackage{multirow}
\usepackage{multicol}
\usepackage{color}
\usepackage{amssymb,amsfonts,amsthm}
\usepackage{amsmath}               
  {
      \theoremstyle{plain}
      \newtheorem{assumption}{Assumption}
  }
\usepackage{algorithm}
\usepackage{algpseudocode}  
\usepackage{graphicx}
\usepackage[misc]{ifsym} 
\usepackage{subfigure} 
\usepackage{subcaption}
\usepackage{textcomp}
\usepackage{xcolor}
\usepackage{dsfont}
\usepackage{mathrsfs}
\usepackage{bbm}
\usepackage{booktabs} 
\usepackage{diagbox}
\usepackage{verbatim}
\usepackage{setspace}
\renewcommand{\qedsymbol}{$\square$}
\newtheorem{lemma}{Lemma}
\newtheorem{theorem}{Theorem}

\newtheorem{definition}{Definition}
\usepackage{etoolbox}

\AtEndEnvironment{theorem}{\hfill\qedsymbol}
\AtEndEnvironment{lemma}{\hfill\qedsymbol}
\AtEndEnvironment{corollary}{\hfill\qedsymbol}
\AtEndEnvironment{definition}{\hfill\qedsymbol}
\AtEndEnvironment{assumption}{\hfill\qedsymbol}
\DeclareMathOperator*{\argmin}{arg\,min}

\allowdisplaybreaks[4]

\title{Stability and Generalization of Adversarial Diffusion Training}
\name{Hesam Hosseini$^*$ \thanks{$^*$This work is performed while the first author was a summer intern at the Adaptive Systems Laboratory at EPFL, Switzerland.} \qquad Ying Cao  \qquad  Ali H. Sayed 
\thanks{
\noindent Emails: seyed.hosseini@epfl.ch, ying.cao@epfl.ch, ali.sayed@epfl.ch\\}
}

\address{School of Engineering, École Polytechnique Fédérale de Lausanne\\
}
\begin{document}
%
\maketitle
\begin{abstract}
Algorithmic stability is an established tool for analyzing generalization. While adversarial training enhances model robustness, it often suffers from robust overfitting and an enlarged generalization gap. Although recent work has established the convergence of adversarial training in decentralized networks, its generalization properties remain unexplored. This work presents a stability-based generalization analysis of adversarial training under the \emph{diffusion strategy} for \emph{convex losses}. We derive a bound showing that the generalization error grows with both the adversarial perturbation strength and the number of training steps, a finding consistent with single-agent case but novel for decentralized settings. Numerical experiments on logistic regression validate these theoretical predictions. 
\end{abstract}
\section{Introduction}
\label{sec:intro}
Machine learning algorithms are designed to fit the training data, as well as generalize to unseen samples, which marks an important distinction between machine learning and a purely optimization-based viewpoint. The difference between empirical and population performance, known as the \emph{generalization gap}, is a central object of study in statistical learning theory \cite{bousquet2002stability,Sayed_2022}.

A well-established framework for understanding generalization is \emph{algorithmic stability}, which relates the sensitivity of an algorithm to perturbations in the training set to its generalization behavior \cite{hardt2016train}. Stable algorithms exhibit small changes in their outputs when a single training example is modified, and this property directly yields generalization bounds.

Adversarial training has emerged as an effective method for enhancing robustness against adversarial examples \cite{ijcai2021p591,a15080283}. However, robust training often suffers from \emph{robust overfitting}, where the generalization gap becomes larger than in standard training \cite{rice2020overfitting,yu2022understanding}. Stability-based analyses have shown that adversarial perturbations degrade stability and enlarge generalization bounds, with the deterioration scaling both with the perturbation radius and the number of training steps \cite{xiao2022stability}.

In parallel, diffusion-based algorithms have become a popular strategy for decentralized learning due to their scalability and communication efficiency \cite{sayed2014adaptation}. Recent work has extended generalization analysis to distributed learning in the clean (non-adversarial) case \cite{bars2023improved}. More recently, adversarial diffusion training was introduced in \cite{cao2025decentralized}, where its convergence and optimization properties were extensively analyzed. However, the generalization behavior of diffusion-based adversarial training remains unexplored.

In this work, we address this gap by providing the first generalization analysis for adversarial training under distributed strategies. While our analysis is restricted to convex loss functions due to space limitations, it offers crucial first insights into this problem. Our specific contributions are:

\begin{enumerate}
\item We derive a stability-based generalization bound for adversarial diffusion training under convex losses. By proving that the algorithm satisfies uniform stability, we show its generalization error grows with the perturbation strength $\epsilon$ and the number of iterations $T$. Our analysis provides a unifying framework, as our bounds seamlessly reduce to known results for single-agent adversarial training \cite{xiao2022stability} and decentralized standard training \cite{bars2023improved} in their respective limits.
\item We combine this stability result with existing optimization guarantees \cite{cao2025decentralized} to characterize the trade-off between optimization and generalization, suggesting practical strategies such as early stopping.
\item We illustrate our theoretical predictions through numerical experiments on logistic regression, confirming the dependence on $\epsilon$ and $T$. Furthermore, our experiments provide new empirical evidence on the influence of network topology.
\end{enumerate}

\section{Problem Setting}

In decentralized learning, multiple agents collaborate to optimize a global objective without centralizing their data. This setup provides benefits in term of scalability and privacy. When agents are exposed to adversarial perturbations, quantifying how these perturbations affect generalization is essential and underexplored. In this work, we formalize this scenario.


We consider a network of $K$ agents connected by a graph topology. The topology is characterized by a doubly stochastic combination matrix $A = [a_{\ell k}]$, where
$a_{\ell k} \geq 0$, and the entries in each column sum to one. Moreover, for any two agents $k$ and $\ell$, if $a_{\ell k} > 0$, there exists a communication link from $\ell$ to $k$.

Each agent $k$ observes independent samples of random variables $(\boldsymbol{x},\boldsymbol{y})$ drawn from a local distribution $\mathcal{D}_k$. Here $\boldsymbol{x}$ plays the role of a feature vector and $\boldsymbol{y}$ plays the role of a label. For a convex loss function $Q_k(w;\boldsymbol{x},\boldsymbol{y})$, the robust local risk is defined as
\begin{equation}
R_k(w) \triangleq\mathbb{E}_{(x,y)\sim\mathcal{D}_k}\!\left[\max_{\delta\in\Delta_\epsilon} Q_k(w;\boldsymbol{x}+\delta,\boldsymbol{y})\right],
\end{equation}
where $\Delta_\epsilon = \{\delta : \|\delta\| \leq \epsilon\}$ represents the set of admissible adversarial perturbations. The inner maximization models worst-case perturbations within a radius $\epsilon$, enabling each agent to train robustly against adversarial examples. The global robust risk aggregates the local risks as
\begin{equation}
R(w) \triangleq \sum_{k=1}^K \pi_k R_k(w), \qquad w^\star \triangleq \argmin_{w \in \mathbb{R}^M} R(w).
\end{equation}
\noindent using weight coefficient $\pi_k \ge 0$ that satisfy $ \sum_{k=1}^K \pi_k = 1.$

In practice, the distributions $\{\mathcal{D}_k\}$ are unknown, and each agent only has access to a finite dataset
\begin{equation}\label{eq:data}
S_k \triangleq \{(x_{k,i},y_{k,i})\}_{i=1}^N, \qquad S \triangleq (S_1,\dots,S_K).
\end{equation}
Each agent is then associated with the following local empirical robust risk
\begin{equation}
\widehat{R}_k(w) \triangleq \frac{1}{N} \sum_{i=1}^N \max_{\delta \in \Delta_\epsilon} Q_k(w;x_{k,i}+\delta,y_{k,i}),
\end{equation}
leading to the global empirical objective
\begin{equation}\label{eq:erm}
R_S(w) \triangleq \sum_{k=1}^K \pi_k \widehat{R}_k(w), \qquad 
\widehat{w} \triangleq \argmin_{w} R_S(w).
\end{equation}
In this work, we focus on using the adversarial diffusion strategy \cite{cao2025decentralized} to solve (\ref{eq:erm}). For simplicity we assume that the inner maximization in $\widehat{R}_k(w)$ has a unique solution, ensuring the adversarial loss
\begin{equation}\label{eq:advloss}
g_k(w;x,y) \triangleq \max_{\delta \in \Delta_\epsilon} Q_k(w;x+\delta,y)
\end{equation}
is differentiable. This holds for common losses like logistic regression under typical perturbation sets \cite{cao2025decentralized}. Agents then update via the adapt-then-combine (ATC) diffusion recursion:
\begin{align}
\phi_{k,n} &= w_{k,n-1} - \mu_n \nabla g_k(w_{k,n-1};x_{k,n},y_{k,n}), \label{eq:adapt}\\
w_{k,n} &= \sum_{\ell \in \mathcal{N}_k} a_{\ell k} \, \phi_{\ell,n}, \label{eq:combine}
\end{align}
with step-size $\mu_n>0$. This recursion incorporates neighbor information while adapting to local observations. For each agent $k$ after $T$ iterations, the algorithm outputs 
\begin{equation}
    F_k(S) \triangleq w_{k,T}
\end{equation}

\noindent The quality of the learned solution is measured by the expected excess risk
\begin{align}\label{eq:risk}
\mathbb{E}_{F,S}[R(F_k(S)) - R(w^\star)]
&\le \underbrace{\mathbb{E}_{F,S}[R(F_k(S)) - R_S(F_k(S))]}_{\epsilon_{\mathrm{gen}}} \nonumber \\
&\quad + \underbrace{\mathbb{E}_{F,S}[R_S(F_k(S)) - R_S(\widehat{w})]}_{\epsilon_{\mathrm{opt}}},
\end{align}

\noindent where the expectation is taken over both the randomness of the data and the algorithm’s internal stochasticity. The term $\epsilon_{\mathrm{gen}}$ quantifies the generalization gap between empirical and population performance, and $\epsilon_{\mathrm{opt}}$ captures suboptimality due to decentralized optimization. While $\epsilon_{\mathrm{opt}}$ has been studied extensively in prior work \cite{cao2025decentralized}, our focus is on bounding $\epsilon_{\mathrm{gen}}$ via algorithmic stability, which we address next.

\section{Stability Analysis}

We now analyze the stability of the adversarial diffusion strategy \eqref{eq:adapt}--\eqref{eq:combine} under convex losses to establish a stability-based generalization bound. Our approach follows a four-step recipe: (i) assume smoothness of local losses, (ii) derive approximate Lipschitz properties of the adversarial objective, (iii) relate stability to generalization via average model stability, and (iv) show that the adversarial diffusion recursion satisfies stability condition.


We adopt standard smoothness assumptions, which are commonly used in the context of decentralized learning and adversarial training \cite{sayed2014adaptation,xiao2022stability,sinha2017certifying,vlaski2021distributed,hardt2016train,cao2025decentralized,bars2023improved}.

\begin{assumption}[\textbf{Smooth loss functions}]\label{assum:smoothness}
For any data point $(x,y)$ and any two model parameters $w,w'$, the loss function $Q_k(w;x,y)$ satisfies:
\begin{equation}\label{eq:lip}
    \|Q_k(w;x,y) - Q_k(w';x,y)\| \leq L_w \|w - w'\|,
\end{equation}
\begin{equation}
    \|\nabla_w Q_k(w;x,y) - \nabla_w Q_k(w';x,y)\| \leq L_{ww} \|w - w'\| ,
\end{equation}
\begin{equation}
 \|\nabla_w Q_k(w;x,y) - \nabla_w Q_k(w;x',y)\| \leq L_{wx} \|x - x'\|.
\end{equation}
\end{assumption}

\noindent These conditions make the loss well-behaved; they are mild and hold for most convex models \cite{xiao2022stability,cao2025decentralized}.


Consider the adversarial loss defined in (\ref{eq:advloss}), then \(g_k\) inherits the convexity and smoothness properties from \(Q_k\) up to a perturbation-dependent term \cite{xiao2022stability,cao2025decentralized}:

\begin{lemma}[\textbf{Convexity of adversarial loss}]\label{lemma:convex}
If \(Q_k(w;x,y)\) is convex in \(w\), then \(g_k(w;x,y)\) is also convex in \(w\).
\end{lemma}

\begin{lemma}[\textbf{Lipschitz properties of adversarial loss}]\label{lemma:adversarial_properties}
Let \(Q_k\) satisfy Assumption~\ref{assum:smoothness}. Then, for all \(w_1,w_2\) and any \((x,y)\), it holds that:
    \begin{align}
    \|g_k(w_1;x,y)-g_k(w_2;x,y)\| \leq L_w\|w_1-w_2\|.
    \end{align}
    \begin{align}\label{eq:AfLip}
         \|\nabla g_k(w_1;x,y) - \nabla g_k(w_2;x,y)\| &\leq L_{ww}\|w_1-w_2\| + 2L_{wx}\epsilon.
    \end{align}
\end{lemma}

\noindent The additive term proportional to \(\epsilon\) in (\ref{eq:AfLip}) captures the sensitivity introduced by adversarial perturbations. We adopt the notion of on-average model stability \cite{lei2020fine} to connect algorithmic stability to generalization.


\begin{definition}[\textbf{On-average model stability}]
Let \(S\) and \(\tilde{S}\) be two independent datasets as defined in (\ref{eq:data}), and let \(S^{(ij)}\) denote \(S\) with the \(i\)-th sample of agent \(j\) replaced by its counterpart in \(\tilde{S}\). For agent $k$ a randomized algorithm \(F_k\) is \emph{on-average \(\eta\)-stable} if
\begin{equation}
\mathbb{E}_{S,\tilde{S},F}\!\left[\frac{1}{KN}\sum_{j=1}^K\sum_{i=1}^N\|F_k(S)-F_k(S^{(ij)})\|\right]\leq \eta.
\end{equation}
\end{definition}

\noindent Intuitively, this notion captures the idea that a stable algorithm should not change its output significantly when a single sample is modified. The smaller the parameter $\eta$, the less sensitive the algorithm is to individual data points, which in turn according to below lemma implies better generalization.

\begin{lemma}[\textbf{Generalization via stability} \cite{lei2020fine}]\label{lemma:gen}
If \(F_k\) is on-average \(\eta\)-stable and (\ref{eq:lip}) is satisfied, then
\begin{equation}
\big|\mathbb{E}_{F,S}[R(F_k(S)) - R_S(F_k(S))]\big| \leq L_w\eta.
\end{equation}
\end{lemma}

\noindent Hence, to bound generalization, it suffices to bound the on-average stability of the adversarial diffusion algorithm.

Recall that the adaptation step is
\begin{equation}
G_{\mu,g}(w) = w - \mu \nabla_w g(w;x,y).
\end{equation}
A key property to guarantee stability is the \emph{non-expansiveness} of the adaptation step: the distance between two iterates \(w_1,w_2\) remains controlled under \(G_{\mu,g}\). It is well-known that SGD is non-expansive under the assumed Lipschitz conditions \cite{hardt2016train} and similarly adversarial updates are approximately non-expansive up to a perturbation-dependent term \cite{xiao2022stability}:

\begin{lemma}[\textbf{Adversarial adaptation update}]\label{lem:sgd_update_props}
Under lemmas \ref{lemma:convex}, \ref{lemma:adversarial_properties}, if \(\mu < 1/L_{ww}\), then for any \(w_1,w_2\), it holds that,
\begin{equation}
\|G_{\mu,g}(w_1) - G_{\mu,g}(w_2)\| \leq \|w_1-w_2\| + 2\mu L_{wx}\epsilon.
\end{equation}
\end{lemma}

\noindent This lemma implies that the adversarial adaptation step approximately preserves stability up to a controlled additive error proportional to \(\epsilon\).

\begin{theorem}[\textbf{Generalization bound for adversarial diffusion training}]\label{thm:adv_diffusion_generalization}
Consider \(K\) agents, each with \(N\) training samples, running the adversarial diffusion algorithm in \eqref{eq:adapt}--\eqref{eq:combine} with step-sizes \(\mu_n \leq 1/L_{ww}\). let \(F_k(S) = w_{k,T}\) be the model parameter of agent \(k\) after \(T\) iterations, then under assumption \ref{assum:smoothness} and Lemmas \ref{lemma:gen} and \ref{lem:sgd_update_props}, it holds for every agent \(k\) that
\begin{equation}
\big|\mathbb{E}[R(F_k(S)) - R_S(F_k(S))]\big|
 \;\leq\; 2L_w\!\left(L_{wx}\epsilon + \frac{L_w}{KN}\right)\sum_{n=1}^T \mu_n .
\end{equation}
\end{theorem}

\noindent This bound highlights two sources of generalization error: (i) the adversarial perturbation bounded by \(\epsilon\), and (ii) the cumulative effect of training iterations in \(\sum_{n=1}^T \mu_n\). In the special case of a fixed step size \(\mu_n = \mu\), the bound simplifies to
\begin{equation}
\big|\mathbb{E}[R(F_k(S)) - R_S(F_k(S))]\big| \;\leq\; 2L_w\mu T\left(L_{wx}\epsilon + \frac{L_w }{KN}\right),
\end{equation}
making the dependence on training duration \(T\) explicit. This result generalizes prior findings in single-agent adversarial training \cite{xiao2022stability} and distributed training without perturbations \cite{bars2023improved} to the adversarial diffusion setting.

\section{Optimization-Generalization Trade-off}


As shown in (\ref{eq:risk}), the performance of a learning algorithm is captured by the expected excess risk, which reflects both optimization and generalization errors. We analyze this trade-off for decentralized adversarial training by combining our generalization results with optimization guarantees.  Theorem~\ref{thm:adv_diffusion_generalization} implies that for a constant step-size $\mu$
\begin{equation}
\epsilon_{\mathrm{gen}} =  O\left(\mu T\left(\epsilon+\frac{1}{KN}\right)\right) \label{eq:our_egen_bound}
\end{equation}
while for the optimization error, the convergence analysis in \cite{cao2025decentralized} yields:
\begin{equation}
\epsilon_{\mathrm{opt}} =O\left(\frac{1}{\mu T}\right) + O\left(\mu\right) \label{eq:opt_error_bound}
\end{equation}

\noindent Combining these bounds gives the overall excess risk:
\begin{align}
\mathbb{E}[R(F_k(S))) - R(w^\star)] = O\left(\mu T\left(\epsilon+\frac{1}{KN}\right)\right) \notag\\ + O\left(\frac{1}{\mu T}\right) +      O\left(\mu\right). \label{eq:total_bound}
\end{align}

\medskip
\noindent
This expression encapsulates the core trade-off in decentralized adversarial training. The $O(\epsilon\mu T)$ term represents the price of robustness, which grows with both the step size and the number of training iterations. The competing $O(\mu T / KN)$ and $O(1 / \mu T)$ terms demonstrate the fundamental optimization-generalization trade-off: longer training improves optimization but harms generalization, an effect mitigated by the total data size $KN$. The $O(\mu)$ term reflects the asymptotic noise controlled by the step size.

This analysis yields  design guidelines: early stopping is necessary to halt training before generalization error dominates; a decaying step-size is useful to balance convergence speed with final error; and collaboration across the network which implicitly increases $N$, directly improves generalization by allowing more training without overfitting.


    
    
    
    

\section{Computer Simulations}

\begin{figure}[!t]
    \centering
    \includegraphics[width=0.44\textwidth]{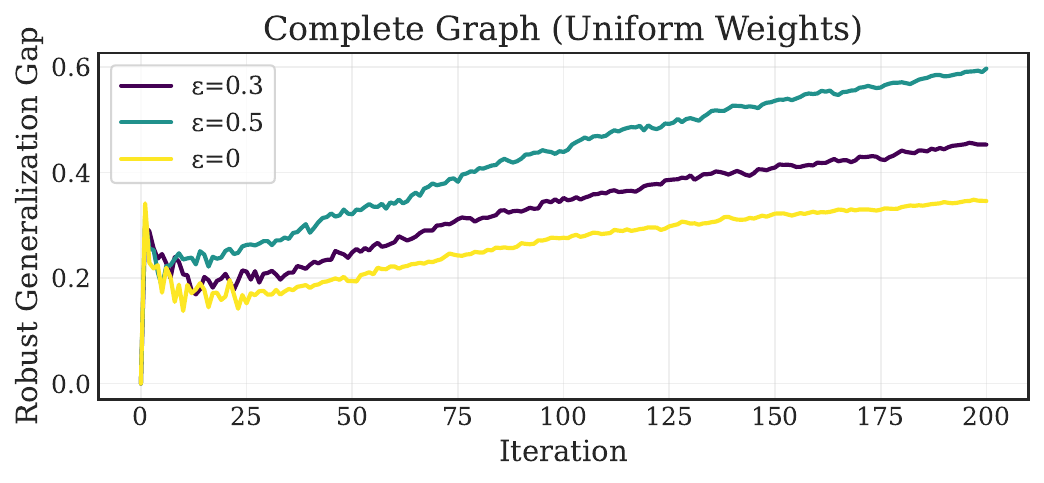}
    \hfill
    \includegraphics[width=0.44\textwidth]{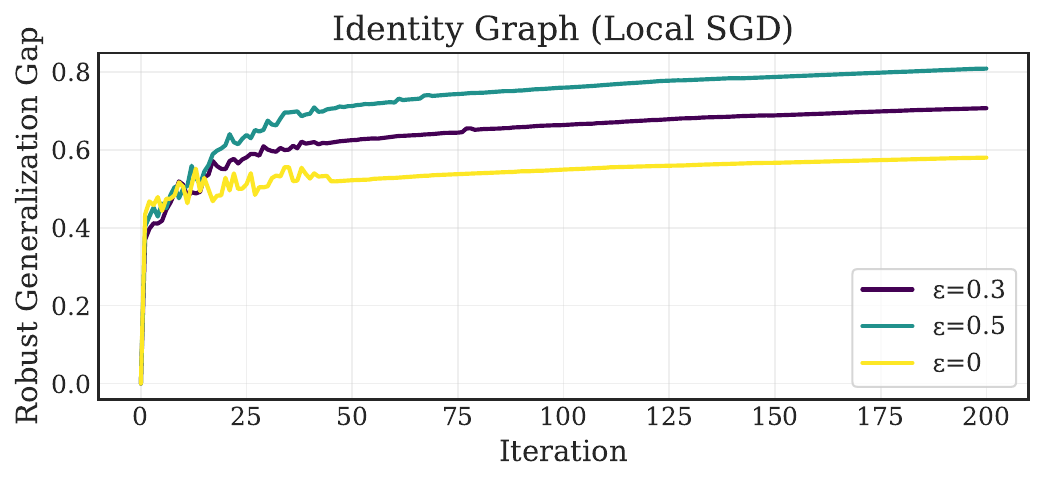}
    
    
    \includegraphics[width=0.44\textwidth]{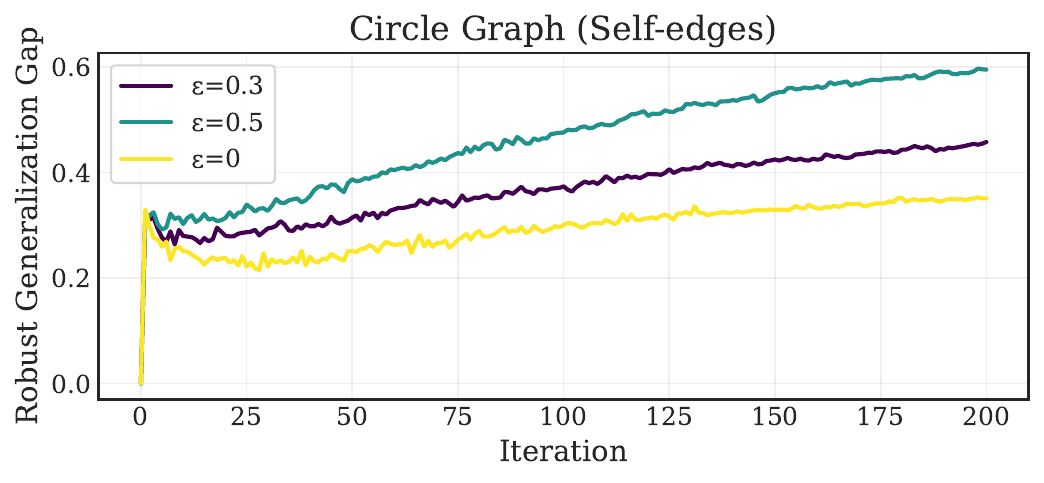}
    \hfill
    \includegraphics[width=0.44\textwidth]{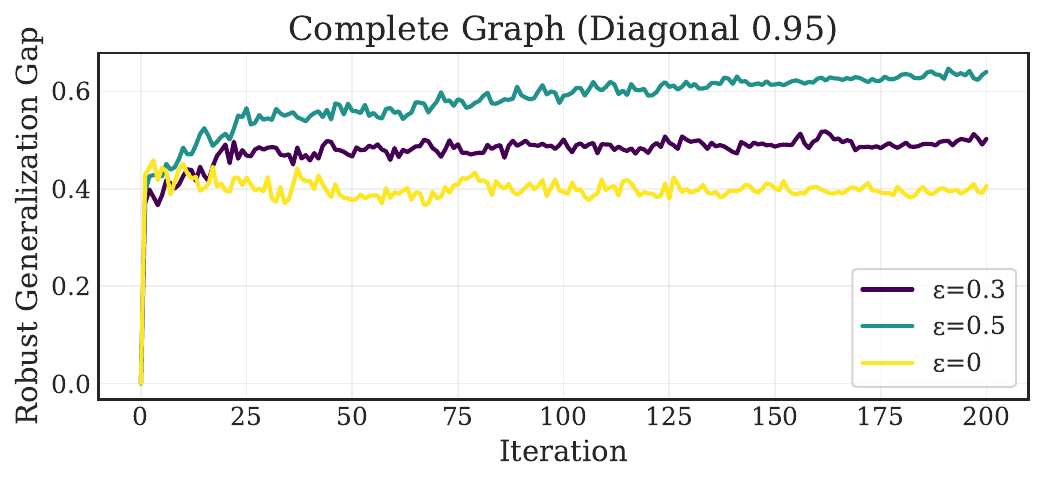}
    
    \caption{Robust generalization gap for different communication graphs.}
    \label{fig:all_gaps}
    \vspace{-0.5cm}
\end{figure}

We illustrate  the theoretical predictions of Theorem~\ref{thm:adv_diffusion_generalization} through logistic regression experiments, a convex setting aligned with our assumptions yet challenging under adversarial training.


 To illustrate the generalization error, we consider a network of $K=10$ agents, each with a local dataset $S_k$ of $N=10$ samples. The robust local risk at agent $k$ is
\begin{align}\label{eq:robust_logistic}
R_k (w) = \mathbb{E}\Big[\max_{\|\delta\|\le \epsilon} \ln(1 + e^{-y(\boldsymbol{x} + \delta)^\top w})\Big],
\end{align}
whose inner maximization under an $\ell_2$-norm constraint admits the closed-form solution via the fast gradient method (FGM) \cite{Sayed_2022,miyato2016adversarial}:
\begin{align}
\delta^\star = - \epsilon \, y \, \frac{w}{\|w\|}.
\end{align}

Synthetic data is generated following \cite{bars2023improved}. Each sample has a label $y \in \{-1,1\}$ (uniformly assigned) and a feature vector $\boldsymbol{x} \in \mathbb{R}^{200}$, drawn from $\mathcal{N}(\mathbf{1}, I)$ for $y=1$ and $\mathcal{N}(-\mathbf{1}, I)$ for $y=-1$, with label flips at probability $0.1$ to avoid linear separability.

We implement the adversarial diffusion algorithm \eqref{eq:adapt}--\eqref{eq:combine} with step-size $\mu=0.03$ and evaluate the robust generalization gap over 15 trials. To study connectivity effects, we test four topologies: (1) Complete graph ($a_{\ell k}=1/K$), (2) Isolated agents ($A=I$), (3) Circle graph ($a_{\ell k}=1/3$ for self and neighbors), and (4) Star-like graph ($a_{kk}=0.95$, $0.05$ for neighbors).

Figure.~\ref{fig:all_gaps} shows the robust generalization gap across topologies as the perturbation radius $\epsilon$ and the number of training iterations $T$ vary. Results confirm Theorem~\ref{thm:adv_diffusion_generalization}: the gap increases with $\epsilon$  and with $T$.

The experimental results also reveal the influence of network topology. While trends with $\epsilon$ and $T$ are almost consistent, the magnitude of the generalization gap depends on the communication graph: well-connected (complete) graphs achieve lower gaps than sparsely isolated agents. Incorporating network connectivity into theoretical bounds is a promising direction for future work.

\section{Conclusion}
We analyzed the generalization performance of the decentralized adversarial training via algorithmic stability analysis and derived a bound for convex loss functions over arbitrary communication graphs. Experiments with logistic regression confirmed that the generalization error increases with perturbation radius and training time, with variations across graph structures. Future work will extend these results to non-convex losses and explore graph design to mitigate robust overfitting.

\bibliographystyle{IEEEbib}
\bibliography{refs}

\appendices
\onecolumn

\end{document}